\documentclass{article}

\usepackage{PRIMEarxiv}

\usepackage[utf8]{inputenc} 
\usepackage[T1]{fontenc}    
\usepackage{hyperref}       
\usepackage{url}            
\usepackage{booktabs}       
\usepackage{amsfonts}       
\usepackage{nicefrac}       
\usepackage{microtype}      
\usepackage{lipsum}
\usepackage{fancyhdr}       
\usepackage{graphicx}       
\graphicspath{{media/}}     

\title{Art Authentication with Vision Transformers}

\author{
  Ludovica Schaerf \\
  Art Recognition AG\\
  Soodmattenstrasse 4\\ 
  CH-8134 Adliswil, Switzerland\\
  \texttt{ludovica.schaerf@gmail.com}
  \And
  Carina Popovici \\
  Art Recognition AG\\
  Soodmattenstrasse 4\\ 
  CH-8134 Adliswil, Switzerland\\
  \texttt{carina@art-recognition.com}
  \And
  Eric Postma \\
  Cognitive Science \& AI \\
  Tilburg University \\
  Tilburg, the Netherlands\\
  \texttt{e.o.postma@tilburguniversity.edu}
}

\begin{document}
\maketitle

\begin{abstract}
In recent years, Transformers, initially developed for language, have been successfully applied to visual tasks. Vision Transformers have been shown to push the state-of-the-art in a wide range of tasks, including image classification, object detection, and semantic segmentation. While ample research has shown promising results in art attribution and art authentication tasks using Convolutional Neural Networks, this paper examines if the superiority of Vision Transformers extends to art authentication, improving, thus, the reliability of computer-based authentication of artworks. Using a carefully compiled dataset of authentic paintings by Vincent van Gogh and two contrast datasets, we compare the art authentication performances of Swin Transformers with those of EfficientNet.
Using a standard contrast set containing imitations and proxies (works by painters with styles closely related to van Gogh), we find that EfficientNet achieves the best performance overall. With a contrast set that only consists of imitations, we find the Swin Transformer to be superior to EfficientNet by achieving an authentication accuracy of over 85\%. These results lead us to conclude that Vision Transformers represent a strong and promising contender in art authentication, particularly in enhancing the computer-based ability to detect artistic imitations.
\end{abstract}


\section{Art attribution and authentication}

Art attribution and art authentication are two significant tasks in the cultural-heritage domain. The former involves identifying the creator of an artwork, while the latter aims to verify whether the artwork was indeed crafted by the presumed artist. These tasks are important because they directly impact the economic and cultural value of artworks~\cite{Spencer2004}. Art attribution entails analyzing various aspects of an artwork, such as its style, materials, and subject, in order to determine the artist responsible for its creation. This can be a challenging task, especially for older works of art where information may be scarce or multiple artists might have been involved. Art authentication involves comprehensive scientific analysis, including the examination of pigments, canvas, paint application techniques, and the historical context~\cite{Sloggett2019}.

\subsection{Computer-based art attribution and authentication} 
\label{sec:history}

With the rise of digital technology, computer-based visual analysis of artworks has provided a new tool to support art attribution and authentication. 
Computer-based methods for art attribution and authentication date back to the turn of the millennium, with the first 'visual stylometry' efforts~\cite{postma2000}. These works are characterized by the development of ad hoc feature extraction methods (including fractal analysis, wavelet coefficients, and edge detection) to represent visual artistic features such as brushstrokes, followed by a machine learning model trained on such features to distinguish the works of the artist from possibly similar works by other artists~\cite{johnson2008image,Qi2013, Liu2016, li2011rhythmic, taylor1999fractal}. 
The excellent pattern-recognition abilities of Convolutional Neural Networks (CNNs) have led to a new wave of studies showing impressive performances on art-classification tasks~\cite{vanNoord2015,vanNoord2017,dobbs2022art} and many other visual tasks~\cite{goodfellow2016deep,Amelio2022,Corradini2022}. These studies involve complex CNN architectures that are trained on large digitized art collections, generally adding to the CNN a last dense (fully-connected) layer. The last layer feeds into a single output neuron in case of art authentication or into $N$ output neurons for art attribution to one of $N$ artists~\cite{cetinic2018fine}.

It should be acknowledged that computer-based art attribution and authentication are not without their limitations and challenges. The first group of limitations stems from the digital nature of the images used in this technique. These images might have deformations and loss of information because of factors such as image resolution, lighting conditions, camera type, and post-processing compression rate. The second group of limitations pertains to connoisseurship. Previous works~\cite{bell2021reflections, zhu2019machine} have discussed the role of the machine as a new type of art expert responsible for attributions and authentications. Bell and Offert (2021)~\cite{bell2021reflections} have highlighted important similarities between human and machine connoisseur approaches, such as knowledge of numerous works by the same artist and related works. However, there are noteworthy differences that constitute limitations of the computer-based techniques. While the computer relies solely on optical information (images), the human connoisseur also considers contextual information, including but not limited to historical knowledge, provenance, and scientific results.

While most of the early studies have primarily focused on traditional machine learning for art-attribution tasks~\cite{Lyu2004,hughes2010quantification,Qi2013,Liu2016}, our paper delves into the more specific task of art authentication, using Vincent van Gogh as a case study. Our paper aims to perform a comparative evaluation of Vision Transformers (ViTs) \cite{Dosovitskiy2021,Liu2021,Liu2022} and CNNs on the art-authentication task, and determine the level of performance that can be attained on this challenging task.

\subsection{Selection of architectures}
\label{sec:stateoftheart}

As we are interested in a comparison between previous state-of-the-art CNN-based methods and ViTs, we have to select representatives of both types of methods. To select a CNN architecture, we determine the best-performing architecture on art-classification tasks by relying on a sample of studies performed over the last 10 years. Although the selected studies have been performed with different methods and datasets, and mostly focused on art attribution instead of art authentication, their performances provide a clear sign of the best-performing architecture.  Table~\ref{table:sota} lists the performances and performance measures for five representative studies over the last 10 years. The performances fall within a limited range,  $78-91\%$. The best-performing study of Table~\ref{table:sota} \cite{dobbs2022art} made use of the ResNet101 architecture~\cite{He2016}.

\begin{table}[ht]
\centering
\caption{Overview of performances in art classification over the last 10 years.}
\label{table:sota}
\begin{tabular}{|l|c|c|c|c|} \hline
    \textbf{ref} & \textbf{year} & \textbf{method}  & \textbf{measure} & \textbf{performance} \\ \hline
    \cite{Qi2013} & 2013 & wavelets  & attribution accuracy & $85-88\%$\\ \hline
    \cite{vanNoord2015} & 2015 & CNN (AlexNet) & mean class accuracy & $78\%$ \\ \hline
    \cite{Liu2016} & 2016 & geometric tight frame &  attribution accuracy & $87-89\%$\\ \hline
    \cite{vanNoord2017} & 2017 & multi-scale CNN & mean class accuracy & $82\%$ \\ \hline
    \cite{dobbs2022art} & 2022 & ResNet101 & mean class accuracy & $91\%$ \\ \hline
\end{tabular}
\end{table}

 Hence, we select ResNet101 as one of the CNN architectures for our experiments. As will be motivated in Section~\ref{subsec:training}, we include another CNN architecture called EfficientNet~\cite{tan2019} in our selection. For the ViTs we will rely on two variants of a state-of-the-art architecture called Swin Transformer~\cite{Liu2021}. 

The outline of the rest of the paper is as follows. Section~\ref{sec:ViT} reviews CNNs and ViTs, highlighting the architectures used in this study, ResNet101, EfficientNet and the Swin Transformer. Section~\ref{sec:exp} details the experimental procedure, and Section~\ref{sec:results} presents the results. Section~\ref{sec:conclusion} ends the paper with a conclusion and discussion of future work.

\section{Convolutional Neural Networks and Vision Transformers}
\label{sec:ViT}

Convolutional Neural Networks gained considerable popularity with the 2012 release of AlexNet, which largely outperformed all previous models at the ILSVRC ImageNet Challenge 2012 \cite{Deng2009, goodfellow2016deep, krizhevsky2017imagenet}. This popularity was further solidified by a continuous stream of improved architectures and layers, most notably InceptionV3, VGG, and ResNet, and current state-of-the-art models such as EfficientNet \cite{tan2019, He2016, szegedy2015rethinking, simonyan2015deep}.

ResNet and EfficientNet are two of the most successful CNNs. ResNet, as described by He et al. \cite{He2016}, is a (potentially) extremely deep CNN. In contrast to a standard CNN, where each stage learns a function $F(x)$ based on input $x$, ResNet stages learn the residual function $F(x) = H(x)-x$ by using skip connections. The use of skip connections allows ResNets to excel due to their increased depth.  EfficientNet represents a class of CNN models introduced by Tan and Le~\cite{tan2019}. These models are optimised by scaling the width, depth, and input resolution of CNNs with a fixed ratio. EfficientNets have demonstrated superior performance compared to ResNets on image classification tasks. In our experiments, we use the variants ResNet-101 and EfficientNetB5. 

Vision transformers are relatively new deep learning architectures that have gained considerable attention and popularity in the computer vision community~\cite{Dosovitskiy2021}. They represent a departure from traditional CNNs by replacing the typical convolutional layers with attention mechanisms~\cite{Vaswani2017}.
In linguistic tasks, the introduction of an attention mechanism facilitated the encoding of long-range contextual information, which led to exceptional results on a wide range of tasks~\cite{Lin2022}. Recent breakthrough performances of GPT4~\cite{openai2023gpt4} and related large language models are due to the power of Transformers. 
One of the main advantages of ViTs is their ability to capture relatively long-range dependencies within an image, which is essential for a wide range of computer vision tasks. This is achieved through the attention mechanism, which allows the model to attend to any region of the image when making predictions, rather than being limited to a fixed image context, like CNNs. ViTs have achieved state-of-the-art results on several image classification benchmarks, including ImageNet, and have shown promising results on other tasks, such as object detection and semantic segmentation. 

\begin{figure}[ht]
\includegraphics[width=\textwidth]{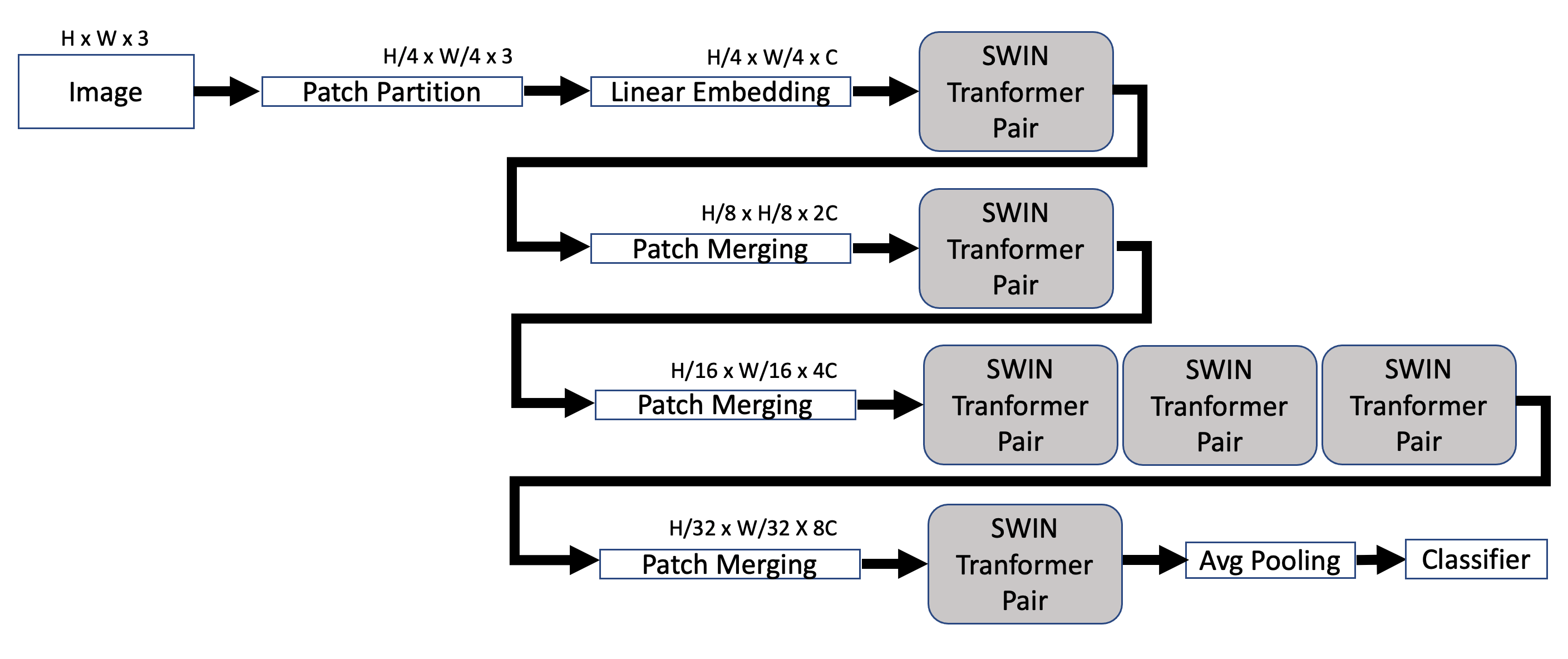}
\caption{Schematic illustration of the "tiny" Swin Transformer (Swin-Tiny). Based on~\cite{Liu2021}.}
\label{fig:swintiny}
\end{figure}

\begin{figure}[ht]
\centerline{
\includegraphics[width=0.5\textwidth]{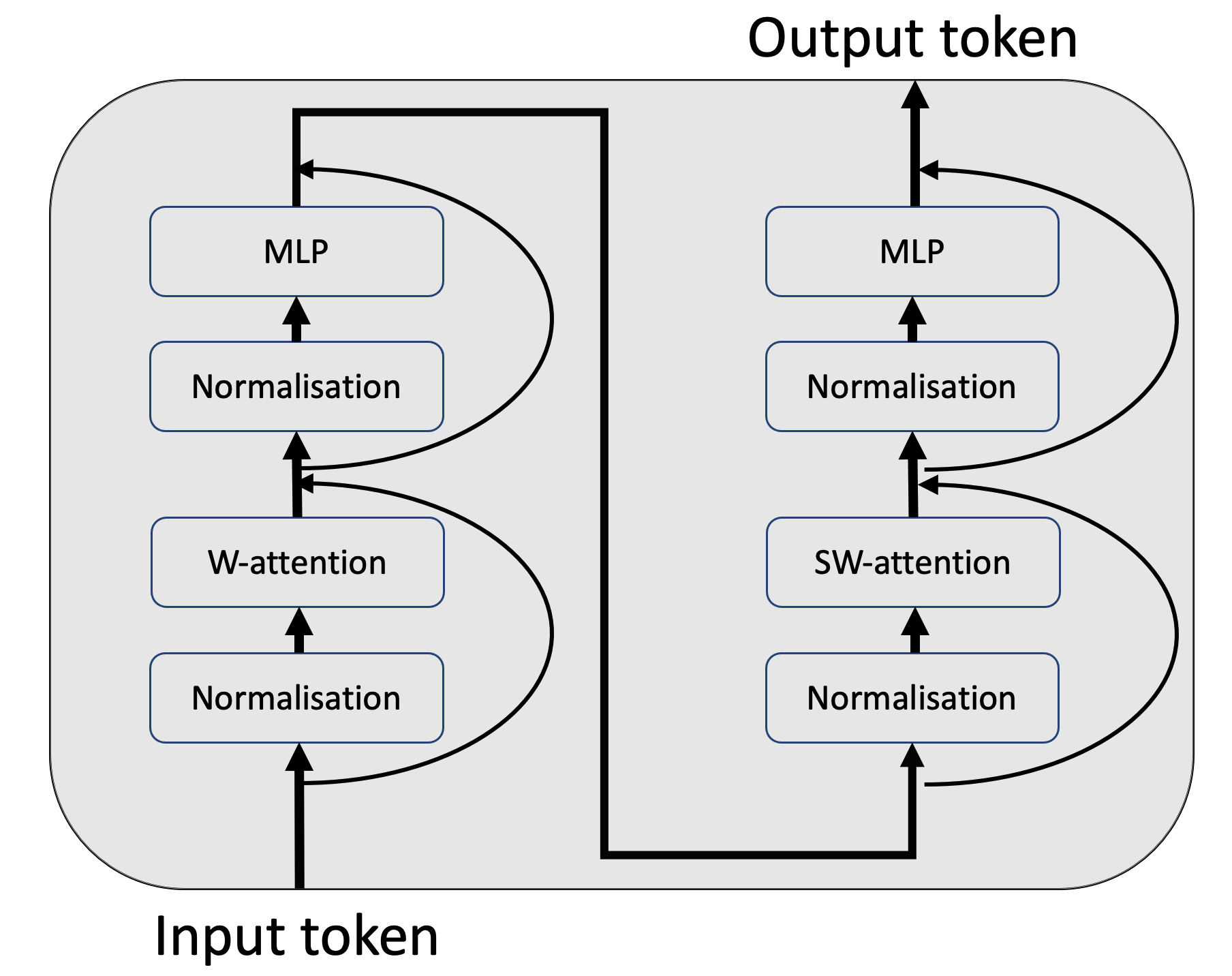}}
\caption{Illustration of the inner structure of the Swin Transformer Pair. Based on~\cite{Liu2021}.}
\label{fig:swinblock}
\end{figure}

The Swin Transformer was recently proposed as a generic Transformer-based backbone for computer vision~\cite{Liu2021,Liu2022}. The basic architecture is hierarchical and employs an efficient self-attention mechanism using shifting windows. Its hierarchical architecture allows for capturing multi-scale relations and its shifting windows mitigate the growth of computational complexity with image size. Figure~\ref{fig:swintiny} illustrates a four-stage Swin Transformer, the so-called "Swin-Tiny" variant. The input comprises an image of size $H \times W \times 3$ which is partitioned into patches of size $W/4 \times H/4 \times 3$ (the rectangle labeled "Patch Partition"). Each patch is embedded into a "token" of size $H/4 \times W/4 \times C$ by means of a linear layer ("Linear Embedding"), where $C$ is an arbitrary dimensionality parameter of the Swin architecture. The token is fed into the building block of the Swin Transformer ("SWIN Transformer Pair"), the inner structure of which is illustrated in Figure~\ref{fig:swinblock}. The first block consist of layer normalisation, multi-head attention, layer normalisation, and two multilayer perceptrons. The multi-head (self-)attention is applied within non-overlapping $M \times M$ windows of the input token ($M = 7$). The curved arrows represent skip connections. The second block is identical to the first, but applies attention to shifted $M \times M$ windows.

To create a hierarchical "pyramid-like" representation, in the second stage of Figure~\ref{fig:swintiny} the "Patch Merging" concatenates all values of each non-overlapping $2 \times 2 \times C$ region into $1 \times 1 \times 4C$ values and uses a linear layer to map these onto $2C$ values. As a result, the $H/4 \times W/4 \times C$ output of the first stage is transformed into a patch-merged output of dimensions $H/8 \times W/8 \times 2C$. The third stage performs the same steps as the second stage, but applies three Swin Transformer Pairs instead of one. Finally, in the fourth stage, patch merging is combined with a single Swin Transformer Pair. In our experiments, the output of the fourth Swin Transformer Pair is Average Pooled and submitted to a binary classifier. 

Apart from the Swin-Tiny variant, three larger variants, which differ from Swin Tiny in the value of the dimensionality parameter $C$ and in the number of Swin Pairs in the third stage $N$, have been proposed. In our experiments, we use the Swin-Tiny ($ C = 96$, $N = 3$) and Swin-Base ($ C = 128$, $N = 9$) variants.

\section{Experiments}
\label{sec:exp}

This section specifies the experiments by discussing the van Gogh dataset~(\ref{subsec:vGdata}), the data preparation and augmentation~(\ref{subsec:dataprep}), the specific CNN and Swin architectures and their hyperparameter settings~(\ref{subsec:training}), and our evaluation procedure~(\ref{subsec:eval}).

\subsection{Van Gogh dataset}
\label{subsec:vGdata}

Our dataset for the authentication task was carefully collected and consists of $654$ images of authentic paintings (authentic set) and $669$ or $137$ images of non-authentic ones (depending on the type of contrast set). 
The resolutions of the images of artworks vary from one reproduction to another. In what follows, we outline the authentic set and two versions of the contrast set: the ``standard contrast set'' and the ``refined contrast'' set. As will be described in Section~\ref{sec:results}, the development of the refined contrast is motivated by the results on the standard contrast set, which reveal that art authentication requires a more constrained selection of artworks in the contrast set.
The composition of each contrast set is described below. 

\subsubsection{Authentic set}

When compiling our authentic set, we have used the standard 'La Faille' \textit{Catalogue Raisonné}~\cite{delaFaille1928} as a reference, meaning that all authentic images used for training are recorded there. Moreover, we have removed from the authentic set the images whose authenticity is questioned by contemporary experts. This approach enables us to mitigate the risk of accidentally introducing fake artworks into the original dataset (label noise). The careful crafting of the authentic set distinguishes this work from previous ones, which are usually trained on images downloaded from WikiArt~\cite{Oliveira2021} (a less reliable source as compared to the established~\textit{Catalogue Raisonné}). 

\subsubsection{Contrast set}
As art authentication involves a binary classification task, we carefully compile a second set that serves as a contrast to the authentic works. This secondary set consists of negative examples, i.e., artworks that are not attributed to van Gogh.

\subsubsection*{Standard contrast set}

The standard contrast set features $69$ imitations: $10$ copies by followers of van Gogh such as Vik Muniz, Blanche Derousse and Jamini Roy; $40$ imitations in van Gogh's style; and $21$ known forgeries, including $8$ produced by the famous forger Wacker~\cite{nelson2011underneath,feilchenfeldt1989van}. In addition, to achieve a balance with the authentic set, the standard contrast set also incorporates $600$ proxies which are paintings by contemporary artists who utilized techniques and styles similar to those of van Gogh -- mainly Post-Impressionism, Cloisonnism, and Japonism. The main proxy artists are Paul Cézanne ($114$ images), Henri de Toulouse-Lautrec ($48$ images), Maurice Prendergast ($47$ images), and Henri Matisse ($47$ images).

\subsubsection*{Refined contrast set}

Including proxies in the standard contrast set introduces painting styles that differ greatly from those of van Gogh.  Hence, for the construction of our refined contrast set, we remove all proxies and gather additional imitations from auction archives.  We include $68$ additional images that were cataloged as being inspired by van Gogh: 50 images are described as \textit{After Vincent van Gogh}, 14 are in \textit{Manner of Vincent van Gogh}, 2 are \textit{Attributed to Vincent van Gogh}, 1 is \textit{Circle Vincent van Gogh} and 1 is \textit{Follower Vincent van Gogh}. Table~\ref{table:compositiondetail} shows the composition of the refined contrast set relative to the standard contrast set.

 \begin{table}[ht!]
\caption{Composition of the standard and refined contrast sets. The proxies for the standard contrast set consist of artists that painted in the same styles as van Gogh, whereas for the refined contrast set the imitations are expanded and include imitations from auction records.}
\label{table:compositiondetail}
\centering
\begin{tabular}{|l||c|c|} \hline
      & \textbf{standard contrast set} & \textbf{refined contrast set} \\ 
\textbf{type} & \textbf{\# images} & \textbf{\# images}\\ \hline \hline
\textbf{imitations}  & 69 & 137 \\ \hline
\textbf{proxies}  & 600 & 0 \\ \hline
\end{tabular}
\end{table}

\subsection{Data preparation}
\label{subsec:dataprep}

\begin{figure}[ht]
\includegraphics[width=0.49\textwidth, height=6cm, angle=0]{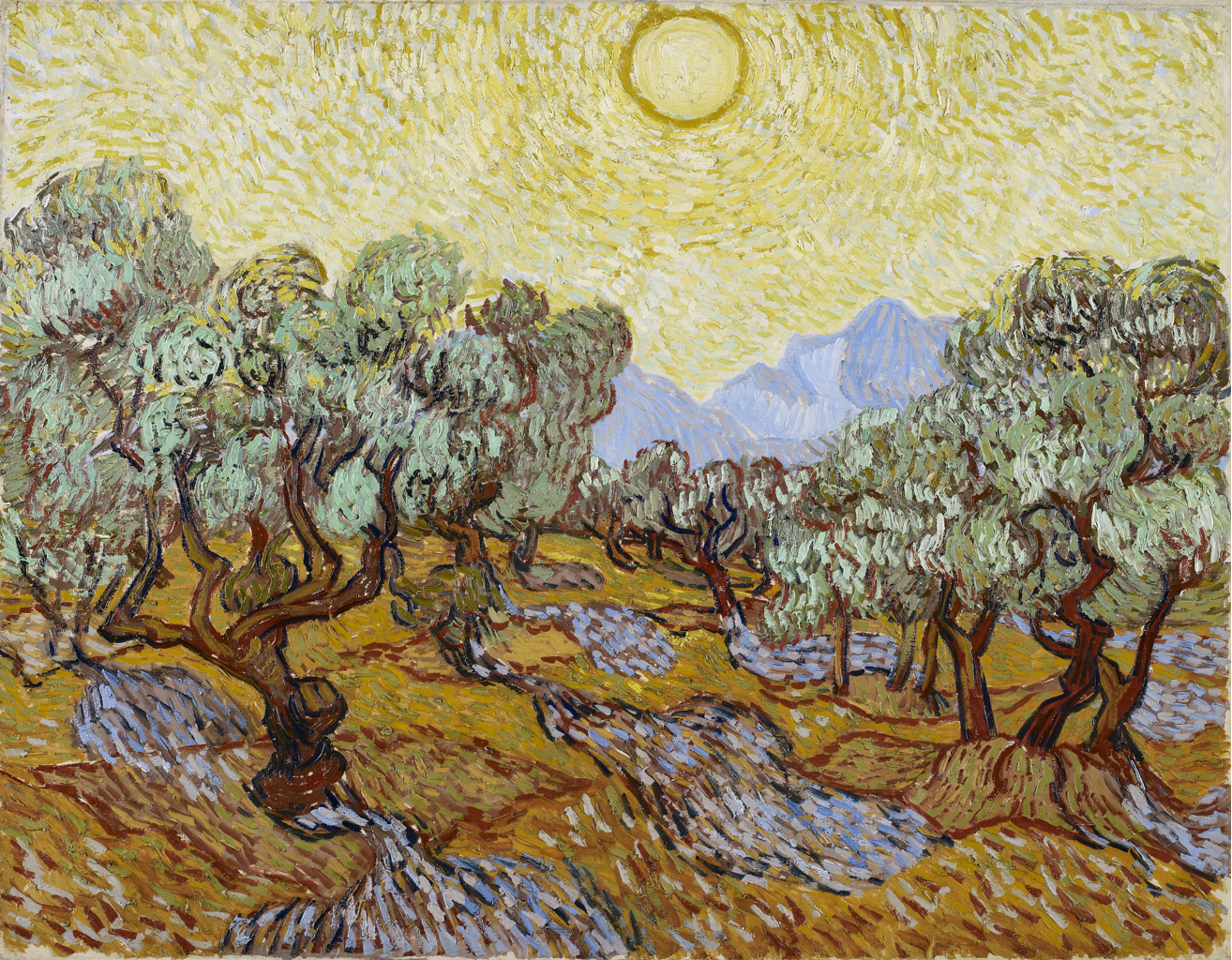}
\includegraphics[width=0.49\textwidth, height=6cm, angle=0]{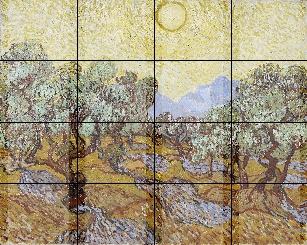}
\caption{Left side: original image. Right side: $4 \times 4$ grid and center cropped squared patches highlighted in bright regions. The image shows the \textit{Oliviers avec ciel jaune et soleil}. Collection: Minneapolis Institute of Art, Location: Minneapolis Institute of Art, oil on canvas. Availability: public domain, CR number: F710.}
\label{patchesdiagram}
\end{figure}

The dataset consists of sub-images of paintings, i.e., RGB images normalized to a fixed size of $256 \times 256$ pixels, and the channel values normalized to the unit interval. The sub-images are created by dividing the whole image into $2^p \times 2^p$ equally sized units, with $p$ depending on the resolution of the original image as follows: $p=2$, if the smaller side of an image is larger than $1024$ pixels, and $p=1$, if the smaller side is larger than $512$ pixels and smaller than $1024$. For all images, regardless of the resolution, we also include the sub-image of the center-cropped square stemming from the full image. Figure~\ref{patchesdiagram} exemplifies the generation of 16 squared, center-cropped patches from an authentic van Gogh painting. This patching method allows the models to extract very fine-grained brushstroke level information from the smaller patches, but also more compositional and representational features from the full patch and the larger patches. Some of the examined architectures require an input size of $224 \times 224$ pixels. In that case the original $256 \times 256$ sub-images were downsampled using bicubic resampling.

To emphasize the importance of imitations over proxies, in the standard contrast set, we assign sample weights $w_{im}$ to the imitations. In preliminary experiments, we found that $w_{im} = 10$, showing that imitations weight ten times more than proxies, yields the best results. This value will remain consistent across the experiments conducted using the standard contrast set described in this study. In the refined contrast set, we did not employ sample weighting, setting $w_{im}=1$.

We evaluate each model in $N=20$ experiments. In each experiment, we randomly assign the paintings including their constituent patches, to the training, validation, and test partitions. These random assignments result in $N$ training, validation, and test partitions. Each model is trained and evaluated on exactly the same $N$ partitions. This ensures that each architecture is trained and evaluated in the same manner, which enables a fair comparative evaluation. Table~\ref{table:partitions} lists the compositions of the partitions in terms of the number of images for the authentic and contrast sets. In each experiment, a randomly selected subset of authentic images of approximately the same size as the size of contrast images is used for training.  

\begin{table}[ht]
\caption{Number of images in each of the three partitions for the experiments with the standard and refined contrast set.}
\centering
\begin{tabular}{|l||c|c|c||c|c|c|} \hline
 & \multicolumn{3}{|c||}{\textbf{standard contrast}} & \multicolumn{3}{|c|}{\textbf{refined contrast}}\\
\hline
 \textbf{\# images} & \textbf{training} & \textbf{validation} & \textbf{test} & \textbf{training} & \textbf{validation} & \textbf{test} \\ \hline \hline
\textbf{authentic set} & 520 & 78 & 73 &  87 &  20 & 30\\ \hline
\textbf{contrast set} & 523 & 65 & 65 & 87 & 20 & 30 \\ \hline
\end{tabular}
\label{table:partitions}
\end{table}
 
Because we subdivide each image into patches, the actual number of patches in each partition is much larger. For instance, for the experiments with the standard contrast set, the actual numbers vary slightly (because images differ in their number of patches): About fifteen thousand patches in the training set and two thousand patches in the validation and test partition each. We emphasize that all patches of each painting are always assigned to the same partition. As a consequence, the test set always consists of patches that were not part of the training or validation partitions. 

\subsection{Architectures and training procedure}
\label{subsec:training}

The recent outstanding art-classification results reported by Dobbs and Ras~\cite{dobbs2022art}, as discussed in Section~\ref{sec:stateoftheart}, has led us to choose ResNet101, the 101-layer version of ResNet~\cite{He2016}, as a representative CNN for our van Gogh authentication task. Although ResNet101 represents the state-of-the-art in art classification, it may not be the most robust CNN available. Therefore, to provide a more comprehensive evaluation, we include another CNN in our analysis that better represents the class of modern CNNs: EfficientNet~\cite{tan2019}. Specifically, we select EfficientNetB5, as its complexity (measured by the number of parameters) roughly matches that of the simplest Swin Transformer. 
For our experiments, we utilize two variations of Swin Transformers (Swin Tiny and Swin Base), with the detailed description of the Swin Transformer architecture provided in Section~\ref{sec:ViT}.

Using the standard contrast set, the four architectures examined are: EfficientNetB5, ResNet101, Swin-Tiny and a larger version called Swin-Base. The latter is included to determine the potential beneficial effect of this larger Swin Transformer variant. EfficientNetB5 has 28M parameters, ResNet101 44.7M parameters, Swin-Tiny and Swin-Base have 28M and 88M parameters, respectively.  
All  architectures are pretrained on ImageNet~\cite{Deng2009}. ResNet101, EfficientNetB5, and Swin-Tiny are pre-trained on the 1K version of ImageNet, whereas Swin-Base is pre-trained on the 22K version of ImageNet. 

In preliminary experiments we explored three variants of transfer learning: (i) freezing the base architecture and training a new top layer (the standard method of transfer learning), (ii) initially freezing the base, training the new top layer, and subsequently training the base and top with a small learning rate, and (iii) unfreezing all layers and training the entire architecture with a small learning rate. It turned out that variant (iii) gave the best results for all architectures, which is in line with previous findings in art classification \cite{gonthier2021analysis, cetinic2018fine}. Hence, in contrast to what is typical to transfer learning, we employed variant (iii), where the top was defined as a randomly-initialized dense layer. For the initialization, we use ``He normal'' initialization~\cite{He2015} that ensures that the random weight values do not saturate the receiving neurons' activations. To this end, the $w_n$ values of the weights feeding into a neuron are drawn from a (truncated) normal distribution with $\mu = 0$ and $\sigma = \sqrt(2/w_n)$. For all architectures, training is performed with binary cross-entropy as loss function, the Adam optimizer, batch size $32$, learning rate $0.0001$, early stopping (patience $= 20$ epochs and minimum delta $= 0.001$), and imitation-sample weights $w_{im}=10$.

For the experiments with the refined contrast set, we apply the same training procedure but do not use imitation-sample weights and restrict ourselves to EfficientNetB5 and Swin-Tiny. The motivation for focusing on these two architectures is twofold: (i) both architectures perform best in the experiments with the standard contrast set, and (ii) comparing the performances of these architectures is fair due to their almost equal parameter complexity. 

\subsection{Evaluation procedure}
\label{subsec:eval}

For each architecture, we performed $N=20$ experiments, and report the average prediction accuracies for individual patches and for the entire paintings. The latter is determined for each artwork by taking the mean of the predictions of its constituent patches, including the sub-image with a center-cropped square stemming from the full image. 
To further understand the model's performance, we present accuracies per class, distinguishing between the authentic and the contrast classes. Additionally, within the contrast set, we provide separate accuracies for proxies and imitations.
\section{Results}
\label{sec:results}

In this section, we present separately the results for the experiments conducted with both the standard contrast set and the refined contrast set.

\subsection{Results for the standard contrast set}

Table~\ref{table:results} reports the results obtained with the standard contrast set. For each of the examined architectures (with the pretraining variants mentioned in Section~\ref{subsec:training}), it lists the mean accuracy for the patches and the entire paintings, as well as  the number of parameters for each architecture. 

\begin{table}[ht]
\centering
\caption{Overview of the results obtained with the standard contrast set. For each architecture, the accuracies averaged over $N=20$ experiments and standard deviations are listed for the patches and entire images. }
\begin{tabular}
{|l|r||c|c|}
\hline
           & \textbf{number of} &   \textbf{patches} & \textbf{paintings} 
        \\ 
\textbf{architecture }
& \textbf{
parameters}  
& \textbf{accuracy (SD)} & \textbf{accuracy (SD)} \\ \hline \hline
EfficientNetB5  & 30M & {\bf 0.912} (0.027) & {\bf 0.925} (0.032)  \\ \hline
ResNet101   & 45M & 0.861 (0.027) & 0.868 (0.032) \\ \hline
Swin-Tiny   & 28M   & 0.893 (0.017) &  0.895 (0.023) \\ \hline
Swin-Base  & 88M   &  0.898 (0.018) & 0.894 (0.024) \\ \hline
\end{tabular}
\label{table:results}
\end{table}

From these results, we draw three observations. The main observation is that EfficientNetB5 yields the best art-authentication performance, both on patches and on entire images. We reiterate that in terms of the number of parameters, EfficientNetB5 has roughly the same complexity and initialization as Swin-Tiny (i.e., 28M and ImageNet 1K, respectively), which makes it a fair comparison. 
The second observation is that both Swin architectures yield a considerable improvement in performance regarding ResNet101, i.e. accuracies $\approx 0.89-0.90$ roughly matching the performances listed in Table~\ref{table:sota} in Section~\ref{sec:stateoftheart}. The third observation is that although the Swin-Base Transformer performs marginally better than the Swin-Tiny Transformer on patches, it does not result in a better performance on paintings. At first sight, these results suggest  EfficientNetB5 outperforms ResNet101 and the Swin Transformers on art-authentication tasks. However, a closer examination of the results for the constituents of the standard contrast set leads to a different view.

Table~\ref{tab:typepredictions} lists the accuracies for the authentic and standard contrast sets, as well as the two constituent types of contrast artworks: imitations and proxies. The results show the performances obtained by all architectures mainly reflect a successful separation of authentic paintings and artworks by proxies, given that both have accuracies of more than $90\%$. On the other side, the performance on the imitations is considerably lower, despite the use of sample weights. We acknowledge that the task of distinguishing imitations from originals is a much more complex and fine-grained one, than distinguishing proxies from originals. Proxies are artworks created by known artists in their own style, albeit similar to the style of van Gogh, while the imitations (including copies and forgeries) contain only artworks that were created, explicitly or implicitly in the style of van Gogh, with a clear and close emulation of the artist. Thus, this last category contains artworks with a much higher degree of similarity to the authentic ones. 

\begin{table}[ht]
    \centering
    \caption{Painting-based test accuracies for the authentic and standard contrast sets, and the two types of contrast types: imitations and proxies. }
    \begin{tabular}{|l||c|c||c|c|} \hline
                   &  \textbf{accuracy} & \textbf{accuracy} & \textbf{accuracy} & \textbf{accuracy}   \\ 
             \textbf{architecture} &  \textbf{authentic}  & \textbf{contrast} & \textbf{imitations} & \textbf{proxies} \\ \hline \hline
    EfficientNetB5  &   \textbf{0.954} (0.029) & \textbf{0.898} (0.047) & 0.527 (0.32)           & \textbf{0.975} (0.035) \\ \hline
    ResNet101 &  0.905 (0.062)          & 0.836 (0.062)          & 0.446 (0.15)           & 0.917 (0.060) \\ \hline
    Swin-Tiny &  0.912 (0.046)          & 0.880 (0.046)          & 0.526 (0.21)           & 0.954 (0.040) \\ \hline
    Swin-Base &  0.905 (0.047)          & 0.883 (0.057)          & \textbf{0.585} (0.28)  & 0.946 (0.045) \\ \hline
    \end{tabular}
    \label{tab:typepredictions}
\end{table}

Clearly, art authentication requires a fine distinction between imitations and authentic art. 
Hence, the poor performance on the imitations motivated the development of the refined contrast set. The results of our experiments with the refined contrast set are the subject of the next section.

\subsection{Results for the refined contrast set}

As mentioned in Section~\ref{subsec:training}, we trained the two comparable architectures on the van Gogh dataset by using a refined contrast set which only comprises imitations. We did not use sample weights for these experiments. The obtained results are presented in Tables~\ref{tab:typepredictionsforgeries} and \ref{tab:precisionrecall}. 

Table~\ref{tab:typepredictionsforgeries} shows the accuracies for paintings in the authentic and refined contrast sets. We observe that in this case, a much better balance is achieved between the performances on the authentic and contrast artworks. This applies especially to Swin-Tiny, which outperforms EfficientNetB5 and achieves the best overall performance. The much-improved performance on the imitations also suggests the relative improvement of the second dataset with respect to the first, as this second dataset tackles best the core of art authentication: the separation between authentic works and reproductions. Alongside this reasoning, the superior performance of Swin-Tiny on this second dataset suggests a non-negligible improvement over state-of-the-art CNNs.

\begin{table}[ht]
    \centering
    \caption{Painting-based test accuracies for authentic and refined contrast sets, by using the EfficientNetB5 and Swin-Tiny architectures.}

    \begin{tabular}{|l||c|c|} \hline
             & \textbf{painting} & \textbf{painting} \\               & \textbf{accuracy (SD)} & \textbf{accuracy (SD)} \\ 
            \textbf{architecture} & \textbf{authentic} & \textbf{refined contrast} \\ \hline \hline
    EfficientNetB5 &  \textbf{0.956} (0.033) & 0.732 (0.11)   \\ \hline
    Swin-Tiny &  0.875 (0.060) & \textbf{0.842} (0.074)  \\ \hline
    \end{tabular}    \label{tab:typepredictionsforgeries}
\end{table}

Table~\ref{tab:precisionrecall} lists the mean accuracy, precision, and recall for EfficientNetB5 and Swin-Tiny. The latter scores best on all three metrics, showing that Swin-Tiny exhibits the best painting-based authentication performance. 
 
\begin{table}[ht]
    \centering
    \caption{Painting-based accuracies, precision, and recall per type of artwork, with the refined contrast set. Pre-training and number of parameters as indicated in Table~\ref{table:results}. See text for details.}

    \begin{tabular}{|l||c|c|c|} \hline
             & \textbf{ painting} & \textbf{painting} & \textbf{painting} \\ 
            \textbf{architecture} & \textbf{accuracy (SD)}   & \textbf{precision (SD)} & \textbf{recall (SD)} \\ \hline \hline
    EfficientNetB5 & 0.843 (0.051) & 0.804 (0.053)  & 0.756 (0.052)   \\ \hline
    Swin-Tiny &  \textbf{0.858} (0.035) & \textbf{0.818} (0.041) & \textbf{0.802} (0.045)  \\ \hline
    \end{tabular}    \label{tab:precisionrecall}
\end{table}

Table~\ref{table:confusion} provides insight into the degree of overlap of patch predictions made by both architectures. The confusion table shows the percentages of patches predicted correctly and incorrectly by both architectures. While they agree on the majority of correctly-classified patches (79\%), their disagreement is limited to smaller percentages (7.4\% and 5.7\%). Both architectures incorrectly classify a slightly larger percentage (8.2\%) of patches. The differences in correctly predicted artworks suggest that avenues combining the strengths of both models may yield even better performance. In this sense, art authentication may benefit a little from a hybrid CNN-ViT approach that combines the strengths of both architectures. 

\begin{table}[ht]
    \caption{Confusion table showing the percentages of correct and incorrect patch predictions for Swin-Tiny and EfficientNetB5 on the enhanced contrast set.}
    \centering
    \begin{tabular}{|cc||c|c|} \hline 
    & & \textbf{Swin-Tiny correct} & \textbf{Swin-Tiny incorrect} \\ \hline \hline
\textbf{EfficientNetB5} & \textbf{correct}   & 79.0\%  & 5.7\% \\ \hline
\textbf{EfficientNetB5} & \textbf{incorrect} & 7.4\% & 8.2\%  \\ \hline
    \end{tabular}
    \label{table:confusion}
\end{table}

\begin{figure}[h!]   
\centering
  \mbox{\includegraphics[scale=.6]{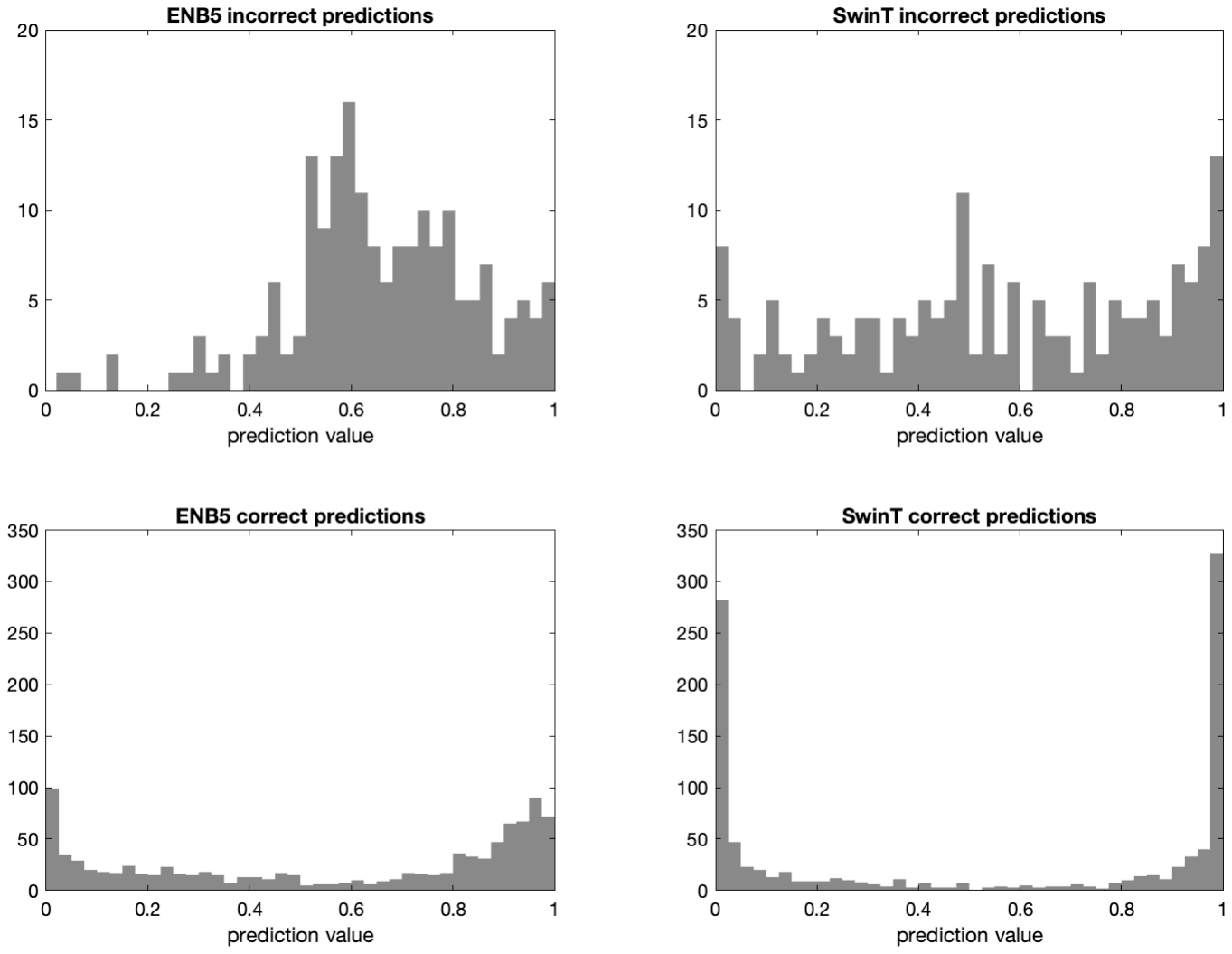}}
\caption{Histograms of the patch predictions for EfficientNetB5 (ENB5, left) and Swin Tiny (SwinT, right). The top rows shows the distribution of incorrect prediction values, the bottom row those of the correct ones. Note that the top and bottom rows have different vertical scales.}
\label{fig:patchpredictions}
\end{figure}

Figure~\ref{fig:patchpredictions} illustrates the differences between both architectures in terms of the distributions of their patch predictions. The histograms for EfficientNetB5 and Swin Tiny are shown in the left and right columns, respectively. The top row displays the incorrect patch predictions, and the bottom row the correct ones. The top left histogram shows a relatively large number of occurrences of wrong predictions in the interval $0.5$-$0.7$ for EfficientNetB5, i.e., the first peak right from the middle. These indicate false positives, revealing a bias toward classifying patches as authentic. Such a peak is not evident for Swin Tiny, although there are more "confident" false predictions at $0$ and $1$ (see the top right histogram). Comparing the bottom two histograms, showing the correct predictions, it is clear that Swin Tiny (right histogram) has a much larger number of very confident predictions (near $0$ and $1$), than EfficientNetB5. These illustrations reveal the subtle ways in which both types of architectures (CNN and Vision Transformer) differ in the realisation of their predictions. To what extent these differences are algorithm-specific is unclear and subject to further investigations. 

\section{Conclusion and future work}
\label{sec:conclusion}

We performed a comparative evaluation of CNNs and Vision Transformers. We found EfficientNetB5 outperforms the Swin-Tiny and Swin-Base Transformers on the standard contrast set, by favoring the classifying of proxies over the classifying of imitations. In our example, this shows that EfficientNetB5 is better able to distinguish between van Gogh and his contemporaries than both Swin Transformers. The Swin-Tiny Transformer was shown to be marginally superior to EfficientNetB5 on a refined contrast set (containing imitations only) that better reflects the essence of art authentication. For the Swin-Tiny Transformer, the change in contrast set was associated with a jump in imitation-classification accuracy from $0.53$ for the standard contrast set to $0.84$ on the refined contrast set. 

While further tests should be carried out to determine the generalizability of these results to other artists' datasets, we also highlight how the deep learning approach to art authentication has an inherent superiority in terms of generalizability to all feature engineering approaches mentioned in Section~\ref{sec:history}, as they require little hyperparameter tuning and do not rely on an isolated feature (i.e. brushstroke) which may not be visible in all artists.

Our results lead us to conclude that visual backbones based on Vision Transformers are at least as viable for art authentication as CNNs and that their predictions largely overlap.  In our future work, we will further explore how Vision Transformers realize their advantage and determine to what extent recently proposed improvements to the Swin Transformer, i.e., the Cross-Shaped Window Transformer~\cite{dong2022}, lead to further improvements on the task of art authentication. 

Future work should also address the limitations arising from the digital nature of the training images. We stress the importance of developing methodologies that can achieve invariance to different camera acquisitions, resolutions, and scales. Additionally, an interesting line of research could explore incorporating contextual information into the models, potentially leveraging multi-modality and textual guidance.

\subsection*{Data availability}

The datasets generated during and/or analysed during the current study are not publicly available due to licensing constraints, but are available from the corresponding author on reasonable request.

\bibliographystyle{unsrt}  
\bibliography{references}

\end{document}